\documentclass[final]{cvpr}

\usepackage{times}
\usepackage{epsfig}
\usepackage{graphicx}
\usepackage{amsmath}
\usepackage{amssymb}

\usepackage[pagebackref=true,breaklinks=true,colorlinks,bookmarks=false]{hyperref}

\usepackage{caption}
\usepackage{subcaption}
\usepackage{mathtools}
\usepackage{array}

\usepackage{bbm}
\usepackage{wrapfig}
\usepackage{diagbox}
\usepackage[misc]{ifsym}

\usepackage{xcolor}
\usepackage{soul}

\definecolor{Highlight}{HTML}{39b54a}
\definecolor{improve}{HTML}{39b54a}
\definecolor{decline}{HTML}{b5394a}

\begin{document}

\title{Deeply Shape-guided Cascade for Instance Segmentation}

\author{Hao Ding$^1$, Siyuan Qiao$^1$, Alan Yuille$^1$, Wei Shen$^{2{(\textrm{\Letter})}}$\\
$^1$Department of Computer Science, Johns Hopkins University\\
$^2$MoE Key Lab of Artificial Intelligence, AI Institute, Shanghai Jiao Tong University\\
{\tt\small \{hding15,siyuan.qiao,ayuille1\}@jhu.edu, wei.shen@sjtu.edu.cn}
\and
}

\maketitle
\let\thefootnote\relax\footnote{$^{\textrm{\Letter}}$ Corresponding Author.}
\begin{abstract}
The key to a successful cascade architecture for precise instance segmentation is to fully leverage the relationship between bounding box detection and mask segmentation across multiple stages. Although modern instance segmentation cascades achieve leading performance, they mainly make use of a unidirectional relationship, i.e., mask segmentation can benefit from iteratively refined bounding box detection. In this paper, we investigate an alternative direction, i.e., how to take the advantage of precise mask segmentation for bounding box detection in a cascade architecture. We propose a Deeply Shape-guided Cascade (DSC) for instance segmentation, which iteratively imposes the shape guidances extracted from mask prediction at the previous stage on bounding box detection at current stage. It forms a bi-directional relationship between the two tasks by introducing three key components: (1) Initial shape guidance: A mask-supervised Region Proposal Network (mPRN) with the ability to generate class-agnostic masks; (2) Explicit shape guidance: A mask-guided region-of-interest (RoI) feature extractor, which employs mask segmentation at previous stage to focus feature extraction at current stage within a region aligned well with the shape of the instance-of-interest rather than a rectangular RoI; (3) Implicit shape guidance: A feature fusion operation which feeds intermediate mask features at previous stage to the bounding box head at current stage. Experimental results show that DSC outperforms the state-of-the-art instance segmentation cascade, Hybrid Task Cascade (HTC), by a large margin and achieves
51.8 box AP and 45.5 mask AP on COCO \texttt{test-dev}. The code is released at: \url{https://github.com/hding2455/DSC}.

\end{abstract}

\section{Introduction}
Instance segmentation~\cite{Ref:ArbelaezPBMM14, Ref:PinheiroCD15, Ref:PinheiroLCD16, Ref:HariharanAGM14, Ref:ChenLY15}, an increasingly active research topic in recent years, is a combination of the elements from two classical computer vision tasks - object detection~\cite{Ref:ErhanSTA14, Ref:Girshick15, Ref:RenNIPS15, Ref:LiuAESRFB16, Ref:RedmonDGF16, Ref:DaiLHS16, Ref:LinDGHHB17,Ref:ZhaoLW18} and semantic segmentation~\cite{Ref:LongSD15, Ref:ChenPKMY18, Ref:JRVSDHT15, Ref:ZhaoSQWJ17, Ref:LuoWLW17,Ref:HarleyDK17}. It is challenging since it requires not only classifying and localizing all the object instances correctly in an image, but also providing a precise segmentation mask for each instance at the same time.

\begin{figure}
\begin{center}
\begin{subfigure}[t]{0.23\textwidth}
     \centering
     \includegraphics[width=1\linewidth]{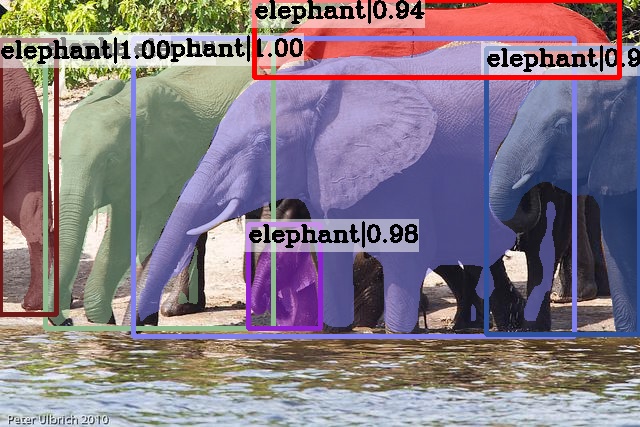}
 \end{subfigure}
 \begin{subfigure}[t]{0.23\textwidth}
    \centering
    \includegraphics[width=1\linewidth]{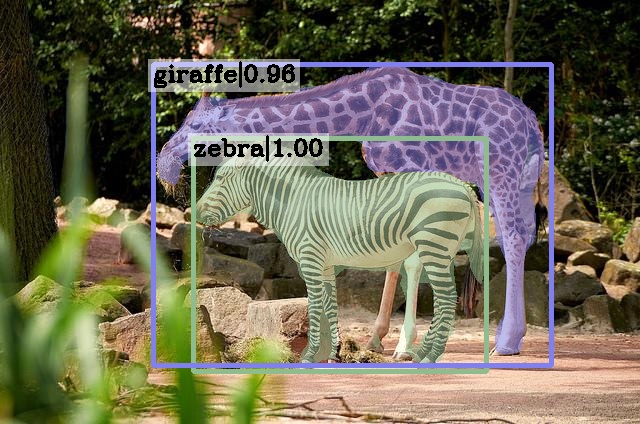}
\end{subfigure}
\end{center}
\caption{Instance segmentation on huddled instances.}
\label{fig:huddle instances}
\end{figure}
To achieve precise instance segmentation, building a cascaded architecture~\cite{Ref:CaiV18} with multi-stage refinement is a promising strategy. As pointed out in~\cite{Ref:ChenPWXLSF0SOLL19}, the key to a successful cascade architecture for precise instance segmentation is to fully leverage the relationship between bounding box detection and mask segmentation across multiple stages. Surprisingly, we find that leading instance segmentation cascades~\cite{Ref:CaiV19,Ref:ChenPWXLSF0SOLL19}, although achieving state-of-the-art performances, mainly make use of an unidirectional relationship, \emph{i.e.}, mask segmentation can benefit from iteratively refined bounding box detection. In this paper, we investigate the opposite direction, \emph{i.e.},  how to take the advantage of precise mask segmentation for bounding box detection. Our aim is to establish a bi-directional relationship between bounding box detection and mask segmentation in a cascade architecture to boost the instance segmentation performance.

Towards this end, we propose a \textbf{D}eeply \textbf{S}hape-guided  \textbf{C}ascade (DSC) for instance segmentation, which iteratively imposes the shape guidances extracted from mask prediction at previous stage on bounding box detection at current stage. DSC plugs the intuitions of Deeply Supervised Nets (DSN)~\cite{Ref:LeeXGZT15,Ref:XieT15}, \emph{i.e.}, 1) enforcing early supervision for intermediate stages and 2) fusing features across multiple stages benefits representation learning, into an instance segmentation cascade. This leads to three key components for shape guidance learning: (1) Initial shape guidance: A mask-supervised Region Proposal Network (mPRN) with the ability to generate class-agnostic masks; (2) Explicit shape guidance: A mask-guided region-of-interest (RoI) feature extractor, based upon the segmented mask at previous stage, focuses on extracting features within a region aligned with the shape of the instance-of-interest rather than a rectangular RoI at current stage; (3) Implicit shape guidance: A feature fusion operation which feeds intermediate mask features at previous stage to the bounding box head at current stage. The shape guidances, either explicitly or implicitly learned from mask-level supervision, are more informative than box-level supervision, and thus are able to help generate more precise bounding boxes.
Then, these bounding boxes can lead to more accurate segmented masks. This forms a positive feedback loop between bounding box detection and mask segmentation across multiple stages in the cascade, facilitating achieving precise instance segmentation.

DSC is easy to implement and can be trained end-to end. Without bells and whistles, on average, it consistently outperforms the state-of-the-art instance segmentation cascade, HTC~\cite{Ref:ChenPWXLSF0SOLL19}, with different backbones by 2.1 box AP and 1.5 mask AP on COCO 2017 \texttt{val} and 1.9 box AP and 1.4 mask AP on COCO 2017 \texttt{test-dev}, thanks to the positive feedback loop between mask prediction and bounding box detection. It is worth mentioning that DSC is good at segmenting huddled
instances, \emph{i.e.}, instances that are crowded together as shown in Fig.\ref{fig:huddle instances}, benefited from the shape guidances. We carefully select a subset of COCO 2017 \texttt{val} which contains a high portion of huddle instances. DSC achieves  significant improvements compared with other methods on this subset.


To sum up, our main contribution is the proposal of a new cascade architecture for precise instance segmentation. It explores a different direction to leverage the relationship between bounding box detection and mask segmentation, forming a positive feedback loop between the two tasks by introducing shape guidances into bounding box detection. It achieves consistent and substantial improvements over the state-of-the-art instance segmentation cascade, Hybrid Task Cascade (HTC)~\cite{Ref:ChenPWXLSF0SOLL19}, on the COCO dataset.


\section{Related Work}
\subsection{Non-cascade Instance Segmentation}
Since instance segmentation combines object detection and semantic segmentation, existing methods for this task can be roughly categorized into two types: segmentation-based and detection-based.
\subsubsection{Segmentation-based Methods}
Segmentation-based methods usually adopt a two-step paradigm - ``segment then identify'', \textit{i.e.}, first perform semantic segmentation to obtain a per-pixel category-level segmentation map for an image, and then identify each object instance therefrom. Liang \emph{et al.}~\cite{Ref:LiangLWSYY18} proposed to identify object instances from the segmentation map by spectral clustering. Kirillov \emph{et al.}~\cite{Ref:KirillovLASR17} partitioned instances from the segmentation map with the help of a learned instance-ware edge map under a MultiCut formulation. Arnab and Torr~\cite{Ref:ArnabT17} made use of the cues from the output of an object detector to identify instances from the segmentation map. Zhang \emph{et al.}~\cite{Ref:ZhangSFU15} predicted instance labels for local patches and merged similar predictions via a Markov Random Field (MRF). They then improved this method by using a densely connected MRF instead, which exploits fast inference~\cite{Ref:ZhangFU16}. Wu \emph{et al.}~\cite{Ref:WuSH16b} proposed a Hough-like Transform to bridge category-level and instance level segmentation, while Bai and Urtasun~\cite{Ref:BaiU17} achieved this by Watershed Transform. There are also some other methods that form instances from a segmentation map by learning an embedding to group similar pixels~\cite{Ref:NewellHD17, Ref:HarleyDK17, Ref:NevenBPG19}. Liu \emph{et al.}~\cite{Ref:LiuJFU17} broke the grouping problem into a series of sub-grouping problems and addressed sequentially.
\subsubsection{Detection-based Methods}
Detection-based methods first generate candidate bounding boxes, then segment the instance mask from each of them. Depending on how to generate the candidate bounding boxes, detection-based methods can be categorized into two classes: anchor-free and anchor-based.
\paragraph{Anchor-free methods.} The early work of anchor-free methods directly used dense sliding-windows as the candidate bounding boxes, such as DeepMask~\cite{Ref:PinheiroCD15}, SharpMask~\cite{Ref:PinheiroLCD16} and InstanceFCN~\cite{Ref:DaiHLR016}, which applied convolutional neural networks to predicting object masks in a dense sliding-window manner. Recent anchor-free methods design more sophisticated to generate mask proposals. YOLACT~\cite{Ref:Bolya19} first learned a dictionary of mask prototypes and then predicted per-instance coefficients to linearly combine prototypes to produce an instance mask. ExtremeNet~\cite{Ref:ZhouZK19} used keypoint detection to predict extreme points, which provide an octagonal approximation for an instance mask. PolarMask~\cite{Ref:Xie2019polarmask} built a polar representation for each instance mask and formulated instance segmentation as instance center classification and dense distance regression in a polar coordinate. TensorMask~\cite{Ref:ChenGHD19} revisited the paradigm of dense sliding window instance segmentation and represented masks by structured 4D tensors over a spatial domain. SOLO~\cite{Ref:SOLO} and its upgraded version~\cite{Ref:SOLOV2} covert instance segmentation into a classification problem by assigning categories to all pixels inside an instance according to its position and scale. Deep Snake~\cite{ref:PengJPLBZ20} and PolyTransform~\cite{Ref:LiangHMXHU20} predicted position offsets w.r.t. the vertices of the polygonal contour of the mask. CondInst~\cite{Ref:TianSC20} estimated the mask head by conditional convolution kernels~\cite{Ref:YangBLN19} to make discriminative mask predictions and eliminate feature alignment. These anchor-free methods mainly focus on real-time performance, and the high precision of the results is not their first priority.
\paragraph{Anchor-based methods.} Anchor-based methods take anchors as references to predict region proposals as the candidate bounding boxes, and then segment each instance mask using the box as a guide~\cite{Ref:LiQDJW17,Ref:ChenHPS0A18}. This paradigm is known as ``detect then segment'', which is currently the dominant paradigm. Mask-RCNN is a representative instantiation of this paradigm, which extended the well-known anchor-based object detector, Faster R-CNN~\cite{Ref:RenNIPS15}, with a mask segmentation branch. Follow-up works, \emph{i.e.}, the variants of Mask-RCNN, improved it by enhancing feature pyramid with accurate localization signals existed in low-level layers~\cite{Ref:LiuQQSJ18}, re-scoring the confidence of a predicted mask to calibrate the misalignment between the mask score and its localization accuracy~\cite{Ref:HuangHGHW19,Ref:ChengWHL20}, or using a more sophisticated bounding box regression method~\cite{Ref:CaoCAKP020}.
\subsection{Cascade Instance Segmentation}
Cascade architectures emerge recently along with the increasing demand for precise object detection. CRAFT~\cite{Ref:YangYLL16} built a cascade structure for both Region Proposal Network~\cite{Ref:RenNIPS15} and Fast R-CNN~\cite{Ref:Girshick15} to get higher quality proposals and detection results. CC-Net~\cite{Ref:OuyangWZW17} excluded easy negative samples at early stages in a cascade. Li \emph{et al.}~\cite{Ref:LiLSBH15} introduced a CNN cascade that operates at multiple resolutions for face detection.

As far as we know, there are only two cascade architectures for instance segmentation, \emph{i.e.}, Cascade Mask R-CNN~\cite{Ref:CaiV19} and Hybrid Task Cascade (HTC)~\cite{Ref:ChenPWXLSF0SOLL19}. As pointed out in~\cite{Ref:ChenPWXLSF0SOLL19}, it is nontrivial to integrate the idea of cascade into instance segmentation. For example, a simple combination of Cascade R-CNN~\cite{Ref:CaiV18} and Mask R-CNN~\cite{Ref:HeGDG17}, \emph{i.e.}, Cascade Mask R-CNN~\cite{Ref:CaiV19}, which iteratively feeds the refined bounding boxes at current stage into next one as high-quality RoIs, only leads to limited gain. HTC~\cite{Ref:ChenPWXLSF0SOLL19} improves Cascade Mask R-CNN by connecting the mask heads at multiple stages through mask information flow.


Both HTC and Cascade Mask R-CNN deliver the message that the mask prediction branch can benefit from the updated bounding box regression. Our method, DSC, shows an orthogonal direction: the object detection branch, \textit{i.e.}, object classification and bounding box regression, can also take advantage of the mask predictions, leading to a positive feedback loop between bounding box detection and mask segmentation.
\section{Methodology}

In this section, we first introduce the overall framework of Deeply Shape-guided Cascade (DSC) for instance segmentation, then elaborate on the three newly introduced key components: the mask-supervised Region Proposal Network (mRPN), the mask-guided ROI feature extractor (shape-guided RoIAlign), and the feature fusion operation. They impose initial, explicit, and implicit shape guidances, respectively, on bounding box detection in the cascade.
\subsection{Overall Framework}\label{sec:overall_framework}

\begin{figure*}
\begin{center}
\centering
\includegraphics[width=1\linewidth]{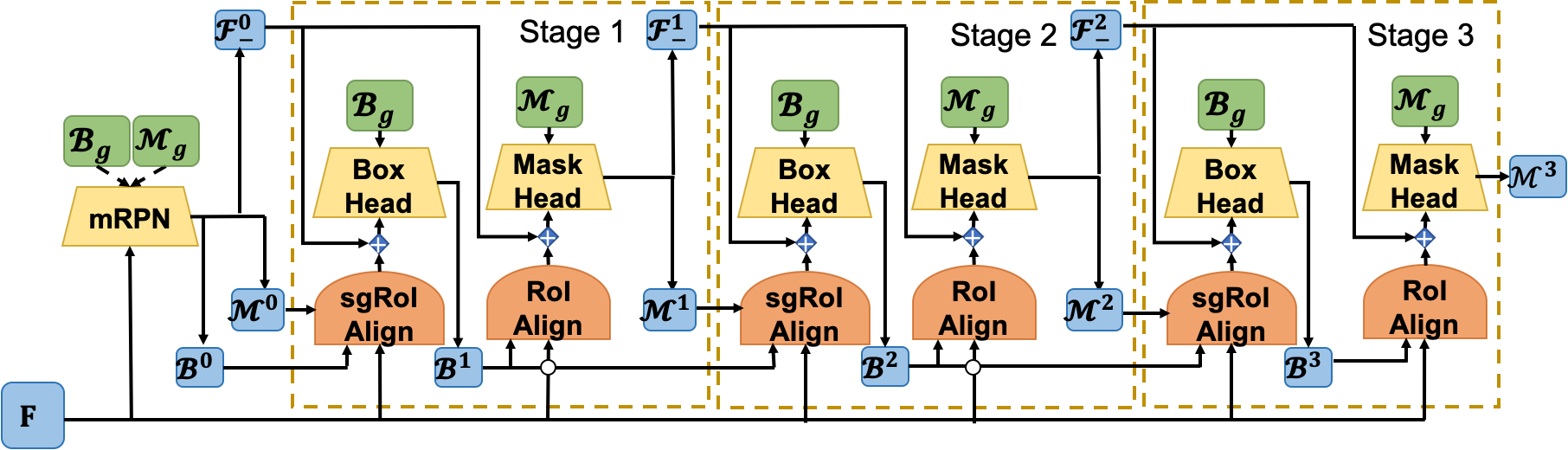}
\end{center}
\caption{The overall framework of Deeply Shape-guided Cascade (DSC) for instance segmentation. Please refer to the first paragraph of Sec.~\ref{sec:overall_framework} for the meanings of the notations.}
\label{fig:overall framework}
\end{figure*}

The overall framework of DSC is shown in Fig.~\ref{fig:overall framework}. It follows the cascade paradigm~\cite{Ref:CaiV19,Ref:ChenPWXLSF0SOLL19}, \emph{i.e.}, first generating a set of instance proposals by a Region Proposal Network (RPN), then iteratively refining the bounding boxes of the proposals and segmenting masks from them by a sequence of box and mask heads. DSC has three new key components:
\begin{itemize}
\item We replace the RPN with the mask-supervised RPN (mRPN), which is guided by both the box supervision $\mathcal{B}_g$ and the mask supervision $\mathcal{M}_g$. Given the feature map $\mathbf{F}$ produced by a CNN backbone as the input, the mRPN produces not only a set of RoIs $\mathcal{B}^0$ but also class-agnostic mask probability maps $\mathcal{M}^0$ corresponding to these RoIs. In addition, it also outputs a set of intermediate mask feature maps $\mathcal{F}_{\_}^0$. Let $\mathbf{B}^0\in\mathcal{B}^0$ be an RoI, then $\mathbf{M}^0\in \mathcal{M}^0$ and $\mathbf{F}_{\_}^0\in\mathcal{F}_{\_}^0$ are its corresponding mask probability map and intermediate mask feature map, respectively. This component involves \textbf{initial shape guidance}, as the shape guidance is learned from the mask supervision and imposed on the early stage (proposal generation stage) of the cascade.
\item For all the box heads in the cascade, we replace the feature extractor, \emph{i.e.}, RoIAlign, with the mask-guided ROI feature extractor, \emph{i.e.}, shape-guided RoIAlign. Shape-guided RoIAlign employs the mask predictions $\mathcal{M}^{t-1}$ at stage $t-1$ to focus feature extraction at stage $t$ within a region aligned well with the shape of the instance-of-interest. This component involves \textbf{explicit shape guidance}, as the learned shape guidance (mask prediction) is directly applied to feature extraction.
\item We conduct a feature fusion operation to enhance the input features of the box head at stage $t$ by integrating them with the intermediate mask feature maps $\mathcal{F}_{\_}^{t-1}$ at stage $t-1$.
This component involves \textbf{implicit shape guidance}, as the intermediate mask features are indirectly learned from the mask supervision.
\end{itemize}
The details of the three components will be described in the following sub-sections.
\subsection{Mask-supervised RPN}

\begin{figure*}
  \centering
    \includegraphics[width=1\linewidth]{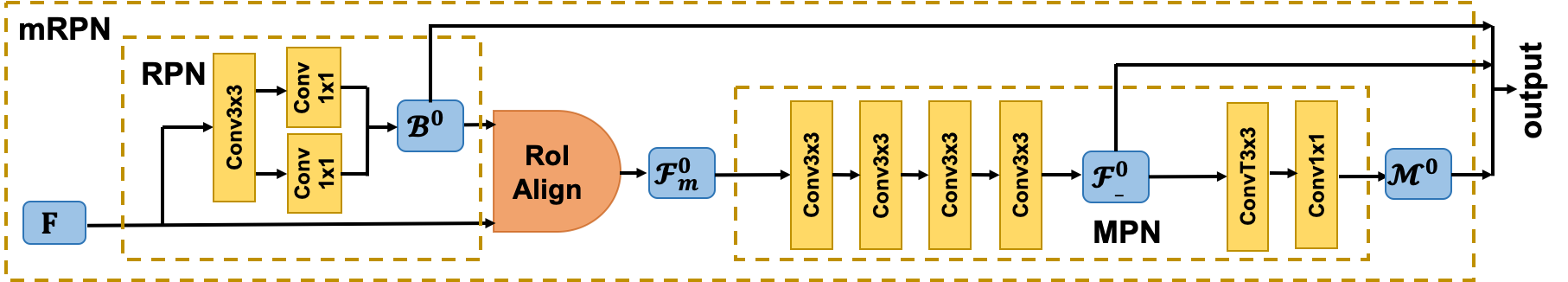}
  \caption{The detail of mask-supervised RPN (mRPN). The supervision flow is omitted for illustration simplicity. }
\label{fig:mRPN_FFB}
\end{figure*}

The detailed design of the mask-supervised RPN (mPRN) is shown in Fig.~\ref{fig:mRPN_FFB}. The mRPN consists of two parts: The first part is the same as the RPN and the second part is a class-agnostic mask generator supervised by the mask supervision $\mathcal{M}_g$, named Mask Proposal Network (MPN). Let $\mathcal{B}^0$ be RoIs outputted by the first part, their feature maps $\mathcal{F}_m^0$ are computed by a RoIAlign layer, which serves as the input of the MPN to generate class-agnostic mask probability maps $\mathcal{M}^0$ within $\mathcal{B}^0$. Both $\mathcal{B}^0$ and $\mathcal{M}^0$ are the outputs of the mPRN, and it also preserves intermediate feature maps $\mathcal{F}_{\_}^0$ of the MPN for further feature fusion.
\subsection{Shape-guided RoIAlign}
Shape-guided RoIAlign (SgRoIAlign) computes feature values under the guidance of the probability of being the instance-of-interest at each location, \emph{e.g.}, a predicted mask probability map $\mathbf{M}$. Similar to RoIAlign, it first divides a RoI $\mathbf{B}$ into $H \times W$ bins. Each bin is denoted by $R_{h,w} = \{(x_h^1, y_w^1), (x_h^2, y_w^2)\}$, where $(x_h^1, y_w^1)$, $(x_h^2, y_w^2)$ are the continuous coordinates of the top-left and bottom-right points of the bin at the $h^{th}$ row and $w^{th}$ column, respectively. $N$ sampling points at continuous location $\{(a_{w,h}^i,b_{w,h}^i)\}_{i=1}^N$ are located uniformly within this bin $R_{w,h}$ for feature extraction. Then, given the $H_p \times W_p$ mask probability map $\mathbf{M}$ aligned with the RoI $\mathbf{B}$, where each element $m(j,k)$ denotes the probability of being the instance-of-interest at a discrete location $(j,k)$, and a sampling point at location $(a_{h,w}^i,b_{h,w}^i)$ on the feature map $\mathbf{F}$, we can compute the corresponding location $(c_{h,w}^i,d_{h,w}^i)$ at probability map $\mathbf{M}$ by equations:
\begin{equation}
\begin{aligned}
c_{h,w}^i &= (a_{h,w}^i - x_1^1) \times \frac{H_p}{H},
~
d_{h,w}^i &= (b_{h,w}^i - y_1^1) \times \frac{W_p}{W}
\label{equation:coordinates_trans}
\end{aligned}
\end{equation}
The feature value $f(a_{h,w}^i,b_{h,w}^i)$ at location $(a_{h,w}^i,b_{h,w}^i)$ on the feature map $\mathbf{F}$ and the probability value $m(c_{h,w}^i,d_{h,w}^i)$ at the corresponding location $(c_{h,w}^i,d_{h,w}^i)$ on the mask probability map $\mathbf{M}$ are computed by  bi-linear interpolation.

Then, the feature representation $f_{\mathbf{B}, \mathbf{M}}(h,w)$ of a bin $R_{h,w}$ is obtained by averaging the multiplications between feature values and probability values plus one at the same sampling points:
\begin{equation}
f_{\mathbf{B}, \mathbf{M}}(h,w) = \sum_{i=1}^{N}\frac{f(a_{h,w}^i,b_{h,w}^i)\times (1 + m(c_{h,w}^i,d_{h,w}^i))}{N}.
\label{equation:Shape_guided_Roi-Align}
\end{equation}

\begin{figure}[h]
  \centering
    \includegraphics[width=0.95\linewidth]{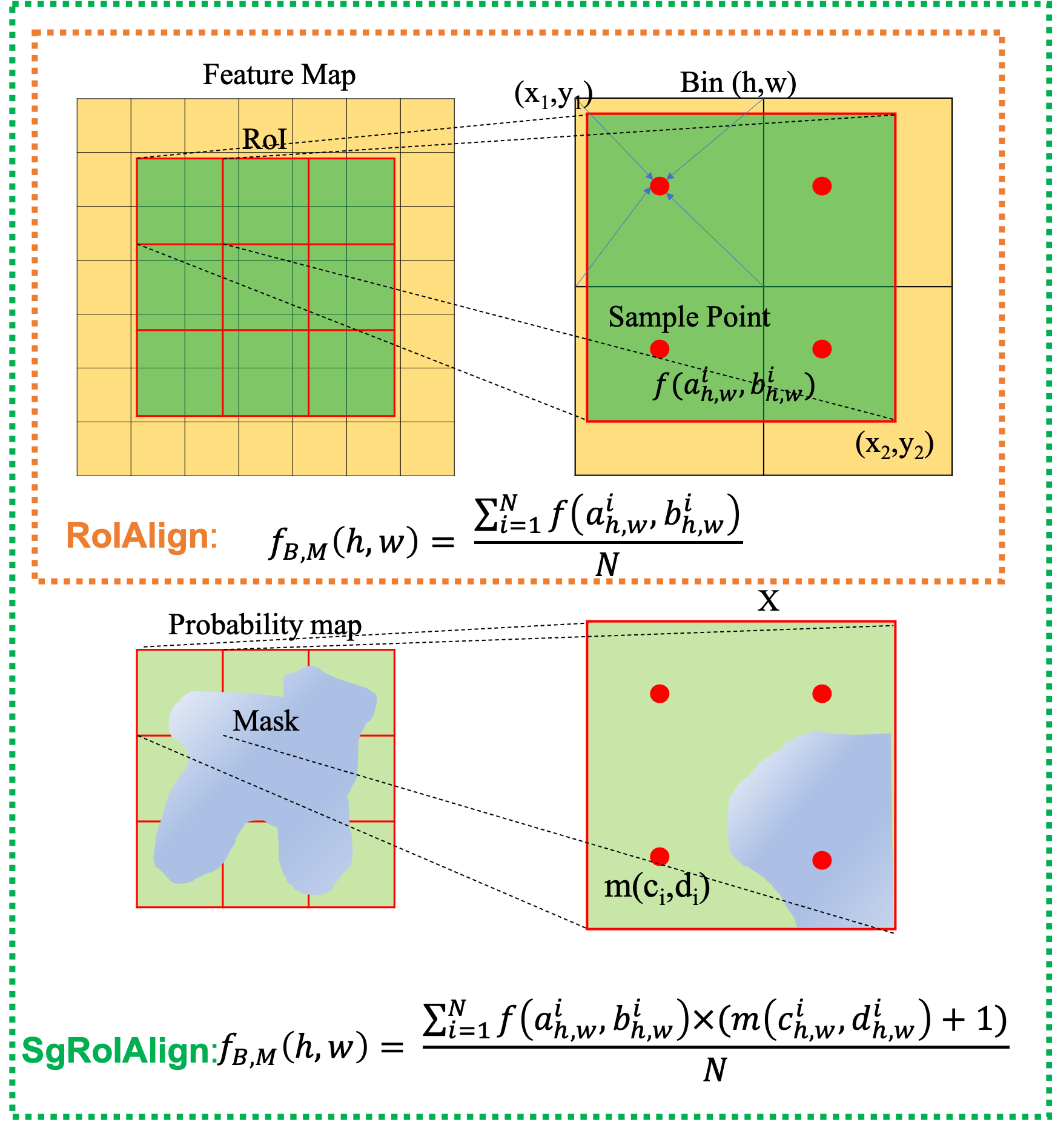}
  \caption{A diagram of shape-guided RoIAlign.}
\label{fig:pooling}
\end{figure}

The intuition of Eq.~\ref{equation:Shape_guided_Roi-Align} is two-fold. On one hand, we want to decrease the impact of the context features and focus feature extraction within the predicted shape region. On the other hand, we do not want to totally exclude the context features, since they are also helpful for object recognition~\cite{OLIVA2007520} and we cannot guarantee that the predicted shape region is perfect. Fig.~\ref{fig:pooling} illustrates a diagram of shape-guided RoIAlign.

Finally, we obtain the small feature map $\mathbf{F}_{\mathbf{B}, \mathbf{M}}$ by repeating the above computation for each bin. We denote this feature extraction procedure, shape-guided RoIAlign, by a function $\mathbf{F}_{\mathbf{B}, \mathbf{M}} = \mathbbm{f}_s(\mathbf{B}, \mathbf{M}, \mathbf{F})$. Correspondingly, the vanilla RoIAlign procedure can be denoted as function $\mathbf{F}_\mathbf{B} = \mathbbm{f}(\mathbf{B}, \mathbf{F})$. Note that in our cascade, as shown in Fig.~\ref{fig:overall framework}, shape-guided RoIAlign and RoIAlign are used to compute the features for box heads and mask heads, respectively. Thus, we rewrite $\mathbf{F}_b=\mathbf{F}_{\mathbf{B}, \mathbf{M}}$ and $\mathbf{F}_m=\mathbf{F}_{\mathbf{B}}$ for notational clearness.

\subsection{Feature Fusion Operation}


The intermediate mask features implicitly learned from the more informative mask supervision at current stage can guide box prediction and mask prediction at next stage. We feed an intermediate feature map $\mathbf{F}_{-}^{t}$ into a 1 $\times$ 1 convolutional layer, then fuse it with the box features $\mathbf{F}_b^{t+1}$ or the mask features $\mathbf{F}_m^{t+1}$ by the element-wise summation. This fusing operation is not trivial since the predicted boxes (RoIs) are iteratively refined in the cascade, which results in feature misalignment among stages.

To eliminate this misalignment issue, recomputing the mask features for the updated RoIs of each stage is a solution, but computationally expensive. The total number of convolutional operations to recompute the mask features is a quadratic function of the total number of the stages.

We propose a simple yet effective strategy for adaptive feature alignment. The basic idea is to reuse the intermediate features of previous stages based on two empirical facts.
(1) Extracting features on the enlarged regions of RoIs with a fixed and relatively small enlargement ratio ($\le 2$) does not degrade the overall performance of the cascade. An enlarged region of an RoI is a rectangular region that shares the same center of the RoI, but the width and height are enlarged by a scale factor $r$.
(2) After bounding box regression, most of RoIs are still in the enlarged regions, and clipping those out-of-region RoIs into the enlarged regions does not degrade the performance of the cascade.

We provide two studies on the cascade Mask R-CNN model to demonstrate these two facts.
For the first one we enlarge RoIs for feature extraction with decreasing enlargement ratios $2^{\frac{3}{3}}, 2^{\frac{2}{3}}, 2^{\frac{1}{3}}$ for stage $1, 2, 3$ of the cascade Mask R-CNN model respectively. This modified cascade Mask R-CNN model achieves competitive performance regarding both box AP (41.3 vs 41.2) and mask AP (36.0 vs 35.9) comparing to the original version.
For the second one we choose different enlargement ratios $2^{\frac{3}{3}}, 2^{\frac{2}{3}}, 2^{\frac{1}{3}}$ to generate enlarged regions for RoIs and calculate the percentage of stay-in-region RoIs, after bounding box regression. We find 98.3\%, 89.16\%, 64.9\% of the RoIs are in the enlarged region after regression. We also clip all out-of-region RoIs and ensure that they are within their corresponding enlarged regions, and observe no AP drop by doing this.

\begin{figure}
  \centering
  \begin{subfigure}[b]{0.215\textwidth}
     \centering
     \includegraphics[width=1\linewidth]{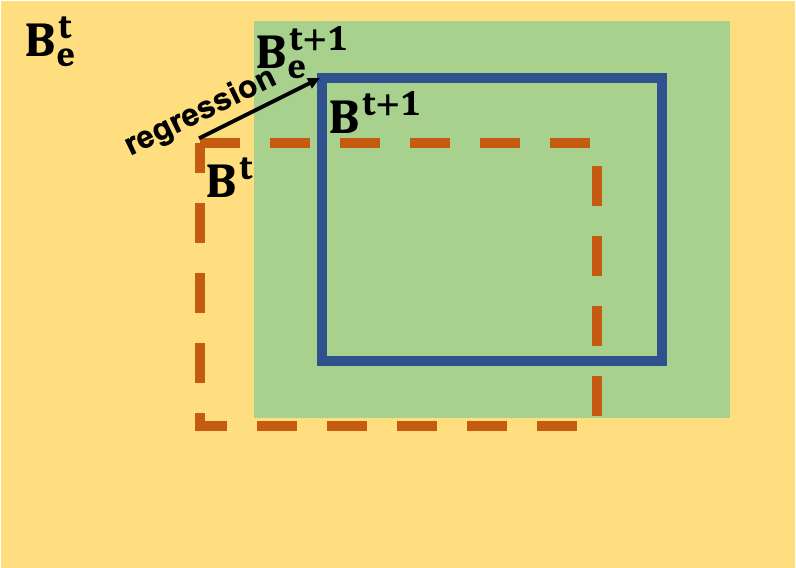}
 \end{subfigure}
 \begin{subfigure}[b]{0.23\textwidth}
    \centering
    \includegraphics[width=1\linewidth]{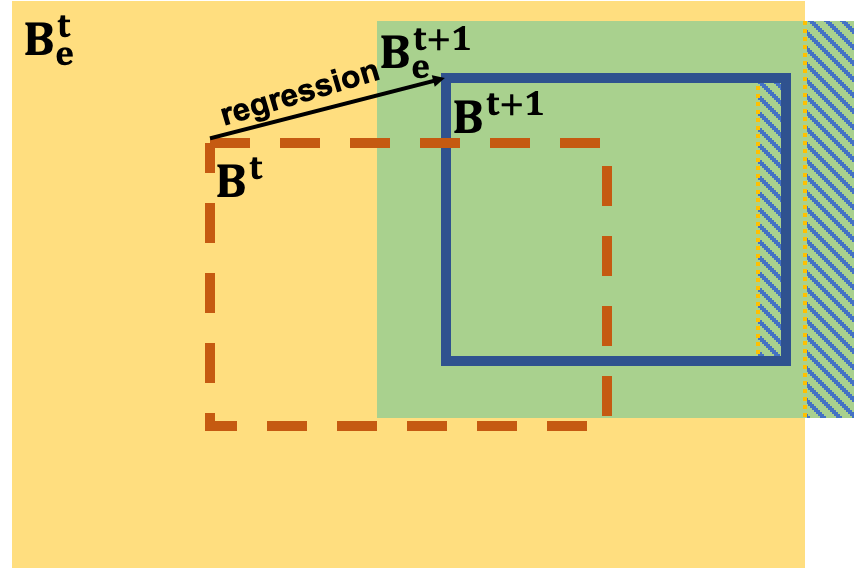}
\end{subfigure}
  \caption{Illustration of the adaptive feature alignment strategy. At stage $t$, feature extraction is performed on the yellow region $\mathbf{B}^{t}_e$, which is an enlarged region of the original RoI $\mathbf{B}^{t}$. At stage $t+1$, $\mathbf{B}^{t}$ is regressed to $\mathbf{B}^{t+1}$. If the enlarged region of $\mathbf{B}^{t+1}$, \emph{i.e.}, the green region $\mathbf{B}^{t+1}_e$, is still within $\mathbf{B}^{t}_e$, we keep $\mathbf{B}^{t+1}$, as shown in the left figure; Otherwise, we clip a part of $\mathbf{B}^{t+1}$ (the shadow area) to ensure its enlarged region is within $\mathbf{B}^{t}_e$, as shown in the right figure.}
\label{fig:Adaptive Feature Alignment strategy}
\end{figure}

The adaptive feature alignment strategy is designed based on these two empirical facts. It consists of two vital steps: 1) RoI enlargement with decreasing enlargement ratios for feature extraction from early to late stages in the cascade; 2) RoI clipping to ensure the enlarged region of each RoI for the following stage is still within the enlarged region of this stage after bounding box regression. Then the features extracted from the enlarged RoI can be reused at next stage without recomputing. Fig.~\ref{fig:Adaptive Feature Alignment strategy} illustrates this strategy. $\mathbf{B}^t$ is an RoI at stage $t$ in the cascade and $\mathbf{B}_e^t$ is its enlarged region. Feature extraction at stage $t$ is performed on $\mathbf{B}_e^t$ and the intermediate mask features are retained. At stage $t+1$, $\mathbf{B}^t$ is regressed to $\mathbf{B}^{t+1}$, whose enlarged region $\mathbf{B}_e^{t+1}$ is guaranteed to be within $\mathbf{B}_e^{t}$ by clipping. Then the mask features for stage $t+1$ can be directly obtained from the retained intermediate mask features at stage $t$. We denote this adaptive feature alignment procedure as a function: $\mathbf{F}^{t+1}_{a-} = \mathbbm{a}(\mathbf{F}^{t}_{-},\mathbf{B}^{t}_e, \mathbf{B}^{t+1}_e)$, where $\mathbf{F}^{t}_{-}$ is the intermediate mask feature map retained at stage $t$ which is not aligned with RoI $\mathbf{B}^{t}$ and the output $\mathbf{F}^{t+1}_{a-}$ is the aligned intermediate mask feature map.



This strategy allows us to extract aligned intermediate features from the retained intermediate features of the previous stage without recomputing. Therefore, the number of convolutional operations for computing the mask features remains a linear function of the total number of the stages.

\subsection{Cascade Pipeline}\label{sec:MGM}

Now we give the formula to summarize the cascade pipeline. At stage $t$, let $\mathbbm{b}^t$ and $\mathbbm{m}^t$ denote the functions of the box head and the mask head, respectively, then we can write the DSC pipeline as:
\begin{equation}
\begin{aligned}
\mathbf{F}^t_m &= \mathbbm{f}(\mathbf{B}_e^t,\mathbf{F}),
\\(\mathbf{M}^t, \mathbf{F}^t_{-}) &= \mathbbm{m}^t(\mathbf{F}^t_m\oplus \mathbf{w}^{t}_m\mathbbm{a}(\mathbf{F}^{t-1}_{-},\mathbf{B}_e^{t-1}, \mathbf{B}_e^t)),
\\ \mathbf{F}^{t+1}_b &= \mathbbm{f}_s(\mathbf{B}_e^t,\mathbf{F}, \mathbf{M}^t),
\\\mathbf{B}^{t+1} &= \mathbbm{b}^t(\mathbf{B}^{t},\mathbf{F}_b^{t+1}\oplus\mathbf{w}^{t+1}_b\mathbf{F}^{t}_{-}),
\label{equation:cascade formulation}
\end{aligned}
\end{equation}
where $\oplus$ denotes the element-wise summation operator, and $\mathbf{w}^t_m$ and $\mathbf{w}^t_b$ are the weights of the 1 $ \times$ 1 convolutional layers for the box head and mask head at stage $t$, respectively, to process the intermediate mask feature map $\mathbf{F}^{t-1}_{-}$. Note that, stage $0$ of the DSC is an mRPN, \emph{i.e.}, $\mathbf{B}^0$, $\mathbf{M}^0$ and $\mathbf{F}^0_{-}$ are produced by the mRPN. Note that, the default resolution of the intermediate mask feature maps is $14\times14$. We can reduce the resolution to $7\times7$ to speed up the cascade. This fast version of DSC is denoted by F-DSC.

\section{Experimental Result}
All experiments are conducted on the COCO 2017 dataset~\cite{Ref:Lin2014MicrosoftCC}, which contains about 118k images with corresponding annotations as the training set and 5k held-out images with annotations as the validation set. The main metric used for evaluation is the standard COCO-style Average Precision (AP) averaged across IoU thresholds from 0.5 to 0.95 with 0.05 as the interval. Both box AP ($\text{AP}_{\text{b}}$) and mask AP ($\text{AP}_{\text{m}}$) are evaluated. We also report $\text{AP}^{50}$ and $\text{AP}^{75}$ (AP at IoU threshold $=0.5$ and $=0.75$). Our models are trained on the 118k training set. Results on the held-out 5k validation set and the 20k test-dev set are reported.
\subsection{Implementation Details}
We use \textit{mmdetection}~\cite{mmdetection} as the code base. For fair comparison. The baselines architectures, \emph{e.g.}, Cascade Mask R-CNN and HTC, are also implemented based on the same codebase.



During training, we sample RoIs once at the first stage and keep the order of the RoIs for every stages to enable using adaptive feature alignment strategy. 
We use SGD as the optimizer with a weight decay of 0.0001 and a momentum of 0.9. The long edge and short edge of each image are resized to 1333 and 800, respectively, without changing the aspect ratio. The scale factors ($r$) to enlarge RoI regions are set to $2^{\frac{3}{3}}, 2^{\frac{2}{3}}, 2^{\frac{1}{3}}$ for stage $1, 2, 3$ respectively.


During inference, following HTC~\cite{Ref:ChenPWXLSF0SOLL19}, at the last box head, the predicted boxes with confidence scores lower than $0.001$ are filtered out. Then, standard non-maximum suppression (NMS) (IoU threshold $=0.5$) is applied to remove duplicated boxes.
\subsection{Benchmarking Results}
All the models are trained for 20 epochs with the learning rate decays at 16th and 19th epochs. For large models that cannot be fitted in the memory, we set the batch size to 8 and adjust the initial learning rate to 0.01.

\begin{figure*}
\begin{center}
\centering
\includegraphics[width=1\linewidth]{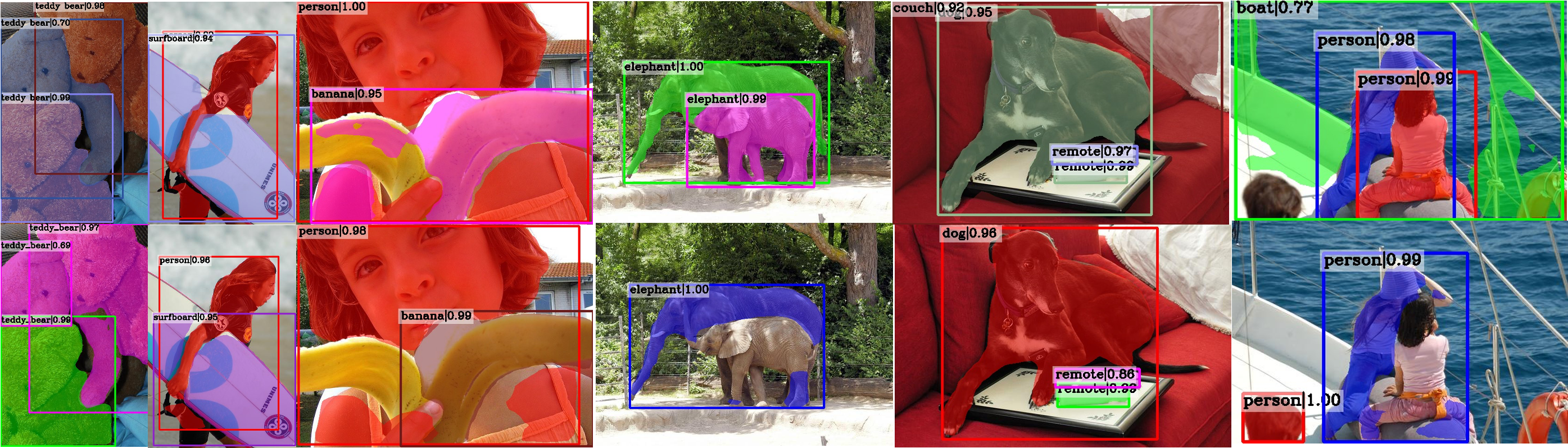}
\end{center}
\caption{Qualitative comparison between DSC (top) and HTC (bottom) on COCO \texttt{val}. All selected images contain huddled instances. HTC is unable to predict object boxes precisely (the left three) or recognize objects correctly (the right three). DSC successfully recognizes all the objects and segments them out.}
\label{fig:qualitative results}
\end{figure*}

{\bf COCO} \texttt{val}: We conduct experiments on COCO \texttt{val} to show the improvements of our DSC over HTC. The experimental results summarized in Table~\ref{tab:DSC validation} show that DSC achieves consistent and remarkable improvements (2.1 box AP and 1.5 mask AP) over HTC with different backbones \emph{e.g.}, ResNet (R) and ResNeXt (X). Note that, DSC always achieves larger improvements under the more strict metric, \emph{i.e.}, $\text{AP}^{75}$, showing high precision of our predictions. Fig.~\ref{fig:qualitative results} shows the qualitative comparisons between HTC and DSC, taking X-101-32x4d FPN as the backbone.

\begin{table*}
\begin{center}
\setlength{\tabcolsep}{0.25em}
\begin{tabular}{|c|c|c|c|c|c|c|c|c|c|}
\hline
Method & Backbone & $\text{AP}_\text{b}$ & $\text{AP}_\text{b}^{50}$& $\text{AP}_\text{b}^{75}$ & $\text{AP}_\text{m}$ & $\text{AP}_\text{m}^{50}$ & $\text{AP}_\text{m}^{75}$ \\
\hline
HTC & R-50 FPN  & 43.3 & 62.2 & 47.1 & 38.3 & 59.3 & 41.4 \\
DSC & R-50 FPN  & {45.8}\color{improve}(+2.5) & {63.4 }\color{improve}(+1.2) & {49.8\color{improve}(+2.7)}& {40.2 }\color{improve}(+1.9) &{61.0}\color{improve}(+1.7) & {43.5}\color{improve}(+2.1) \\
\hline
HTC & R-101 FPN  & 44.8 & 63.3 & 48.8 & 39.6 & 61.0 & 42.8\\
DSC & R-101 FPN  &  {46.6}\color{improve}(+1.8) &  {64.5} \color{improve}(+1.2) & {50.8}\color{improve}(+2.0) & {40.7 }\color{improve}(+1.1) & {62.0} \color{improve}(+1.0)& {44.1}\color{improve}(+1.3)\\
\hline
HTC & X-101-32x4d FPN  & 46.1& 65.3& 50.1& 40.5& 62.5& 43.7\\
DSC & X-101-32x4d FPN  &{48.0}\color{improve}(+1.9) &{65.9}\color{improve}(+0.6) &{52.2}\color{improve}(+2.1) & {42.0}\color{improve}(+1.5) & {63.7}\color{improve}(+1.2)  & {45.6}\color{improve}(+1.9)\\
\hline
\end{tabular}
\end{center}
\caption{Comparison with HTC on COCO \texttt{val}.}
\label{tab:DSC validation}
\end{table*}


{\bf COCO} \texttt{test-dev}: We do a comprehensive comparison with state-of-the-art instance segmentation methods with strong network backbones on COCO \texttt{test-dev}. The results are summarized in Table~\ref{tab:all method}, which show that DSC outperforms these state-of-the-art methods by a larger margin when using the same network backbones. Especially, HTC with deformable convolution (DCN) ~\cite{Ref:DaiQXLZHW17} and multi-scale training (ms train) is a very powerful cascade model for instance segmentation, which already achieved very high performance (50.8 box AP and 44.2 mask AP). Nevertheless, our method, DSC, with the same backbone and training strategy (DCN + ms train) still obtains a significant improvement (1.0 box AP and 1.3 mask AP) and achieves promising results (51.8 box AP and 45.5 mask AP) on COCO \texttt{test-dev}.

\begin{table*}
\begin{center}
\begin{tabular}{|>{\centering\arraybackslash}m{3.7cm}|>{\centering\arraybackslash}m{3.8cm}|>{\centering\arraybackslash}m{0.8cm}|c|c|c|c|c|c|}
\hline
Method & Backbone &Epoch & $\text{AP}_\text{b}$ & $\text{AP}_\text{b}^{50}$ & $\text{AP}_\text{b}^{75}$ &  $\text{AP}_\text{m}$ &$\text{AP}_\text{m}^{50}$ & $\text{AP}_\text{m}^{75}$\\
\hline
\emph{One-stage:} &  &  &  &  &  & &  & \\
BlendMask (ms train)~\cite{Ref:ChenSTSHY20} & R-50+FPN & 36 & -  & -& -& 37.0 &58.9 & 39.7\\
SOLOv1 (ms train)~\cite{Ref:WangKSJL20} & R-50+FPN & 72 & -  & -& -& 36.8 &58.6 &39.0\\
SOLOv2 (ms train)~\cite{Ref:WangZKLS20} & R-50+FPN & 72 & -  & -& -& 38.8 & 59.9 & 41.7\\
CondInst (ms train)~\cite{Ref:TianSC20} & R-50+FPN & 36 & -  & -& -& 38.8 & 60.4&41.5\\
\emph{Two-stage:} &  &  &  &  &  &  &  &\\
FCIS++~\cite{Ref:LiQDJW17} & R-101 & - & - & - & - & 33.6 &54.5 & -\\
MaskLab+~\cite{Ref:ChenHPS0A18} & R-101(JET) & - & - & - & - & 38.1 & 61.1& 40.4\\
PANet~\cite{Ref:LiuQQSJ18} & R-50+FPN & 24 & - & - & -  & 36.6 &58.0 &39.3\\
D2Det~\cite{Ref:CaoCAKP020} & R-101+FPN  & 24 & -  & -& -& 40.2 & 61.5&43.7\\
Mask R-CNN~\cite{Ref:HeGDG17} & R-101+FPN & 24 &41.6 & 62.5 & 45.3& 37.4& 59.5& 40.0\\

MS R-CNN~\cite{Ref:HuangHGHW19} & R-101+FPN &24 & 41.6 &62.3 &46.2 & 38.3 &58.5 &41.5\\
\hline
\emph{Cascade:} &  &  &  &  &  &  &  &\\
Cascade Mask R-CNN\cite{Ref:CaiV19} & R-50+FPN & 20 & 42.8 &61.6 & 46.5 & 37.0 & 58.6 & 39.9\\
HTC~\cite{Ref:ChenPWXLSF0SOLL19} & R-50+FPN & 20 & 43.6 & 62.6& 47.4& 38.5 & 60.1 & 41.7\\
DSC (ours) & R-50+FPN & 20 & 46.0 &63.9 &50.1 & 40.5 & 61.8& 44.1\\
HTC~\cite{Ref:ChenPWXLSF0SOLL19} & R-101+FPN & 20 & 45.1 & 64.2& 49.1& 39.8 & 61.6 & 43.1\\
DSC(ours) & R-101+FPN & 20 & 46.7 & 64.7& 50.9& 40.9 & 62.5& 44.5\\
HTC~\cite{Ref:ChenPWXLSF0SOLL19} & X-101-32x4d+FPN & 20 & 46.4 & 65.8& 50.4& 41.0 & 63.2 & 44.4\\
DSC(ours) & X-101-32x4d+FPN & 20 &48.1& 66.3& 52.4& 42.2& 64.1& 45.8\\
HTC (ms train)~\cite{Ref:ChenPWXLSF0SOLL19} & X-101-64x4d+DCN+FPN & 20 & 50.8 & 70.3&55.2 & 44.2 &67.8 &48.1\\
DSC (ms train) (ours) & X-101-64x4d+DCN+FPN & 20 & {\bf 51.8} & {\bf 70.5}& {\bf 56.7}& {\bf 45.5} & {\bf 68.4} & {\bf 49.7}\\

\hline
\end{tabular}
\end{center}
\caption{Comparison with state-of-the-art methods on COCO \texttt{test-dev}.}
\label{tab:all method}
\end{table*}

\subsection{Ablation Study}
In this section, we conduct ablation studies on COCO \texttt{val} to investigate the efficiency of our method, the contribution of each component we introduced for the cascade architecture. We use R-50 as the backbone and 1$\times$ learning rate schedule for all ablation studies.

\begin{table}
\begin{center}
\begin{tabular}{|c|c|c|c|c|}
\hline
 Methods  & $\text{AP}_\text{b}$ & $\text{AP}_\text{m}$ & Inference Time\\
\hline
HTC & 42.3 & 37.4 & 238ms \\
DSC & 44.8\color{improve}(+2.5) & 39.5\color{improve}(+2.1) & 434ms\color{decline}(+196ms) \\
F-DSC & 44.5\color{improve}(+2.2)& 39.4 \color{improve}(+2.0) & 256ms\color{decline}(+18ms) \\
\hline
\end{tabular}
\end{center}
\caption{Comparison among DSC, F-DSC and HTC}
\label{tab:inftime}
\end{table}
\subsubsection{Precision vs Inference Time}
There is always a trade-off between precision and inference time for a method. Since our method has two versions, DSC and a fast version F-DSC. It is necessary to discuss this trade-off for them. HTC is taken as the baseline for reference.

As the results shown in Tab.~\ref{tab:inftime}, comparing to DSC, F-DSC achieves a comparable performance in terms of both Box AP and Mask AP, \emph{i.e.}, only $0.3$ Box AP and $0.1$ Mask AP drops, while it is much faster than DSC, \emph{i.e.}, reducing the inference time by 178ms. Comparing to HTC, F-DSC achieves significant improvements in both Box AP and Mask AP, \emph{i.e.}, $2.2$ Box AP and $2.0$ Mask AP improvements. Moreover, the additional inference time ($18$ ms) is negligible. Since F-DSC performs excellently in both precision and inference time, we conduct the rest experiments based on F-DSC.

\subsubsection{Contribution of Each Cascade Component}
We conduct an ablation study to verify the contribution of each component we introduce for our cascade architecture, including shape-guided RoIAlign, \emph{i.e.}, explicit shape guidance (ExSG), the ``+1'' term in shape-guided RoIAlign (Plus1), the feature fusion operation, \emph{i.e.},  implicit shape guidance (ImSG), and the adaptive feature alignment (AFA) strategy. We also compare a baseline which recomputes the mask features for the updated RoIs of each stage (ReComp) instead of using the AFA strategy.
Since mRPN only provides the initial shape guidance for the first stage of our cascade architecture, we exclude it from the ablation study.

Table~\ref{tab:AblationModules} summarizes the result of ablation study. We observe that excluding either ExSG or ImSG from F-DSC leads to performance degradation, showing that both explicit and implicit shape guidances provide positive feedback for the cascade architecture. we find that removing the ``+1'' term from ExSG leads to a drop of 0.2 Box AP and 0.2 Mask AP. Regarding that the improvement brought by ExSG is 0.3 Box AP/0.4 Mask AP, the ``+1'' term is important. We also observe that, without the AFA strategy, the performance has a noticeable drop (0.5 Box AP and 0.5 Mask AP). This result shows that the misaligned mask features impose negative effect on the implicit shape guidance and the adaptive feature alignment method is effective to address this problem. Recomputing the mask features for the updated RoIs of each stage can solve this misalignment problem and further improves F-DSC by 0.4 Box AP and 0.4 Mask AP, but takes extra 101ms.


\begin{table}
\begin{center}
\begin{tabular}{|c|c|c|}
\hline
 Method & $\text{AP}_\text{b}$ & $\text{AP}_\text{m}$ \\
\hline
F-DSC &44.5 &  39.5\\
F-DSC - ExSG & 44.2\color{decline}(-0.3) &  39.1\color{decline}(-0.4)\\
F-DSC - ImSG & 43.4\color{decline}(-1.1) & 38.8\color{decline}(-0.7) \\
F-DSC - Plus1 &  44.3\color{decline}(-0.2) & 39.3\color{decline}(-0.2) \\
F-DSC - AFA &  44.0\color{decline}(-0.5) & 39.0\color{decline}(-0.5) \\
F-DSC - AFA + ReComp &  44.9\color{improve}(+0.4) & 39.9\color{improve}(+0.4) \\
\hline
\end{tabular}
\end{center}
\caption{Ablation study to verify the contribution of each component we introduce for our cascade architecture. The symbols ``+'' and ``-'' mean including and excluding a component into and from F-DSC, respectively.}
\label{tab:AblationModules}
\end{table}
\subsection{Quantitative Results on huddle Instances}
\paragraph{Huddled Instance Data Collection.} To validate the advantage of DSC on detecting and segmenting huddled instances, we do comparisons between DSC and HTC on subsets of COCO 2017 \texttt{val} with different proportions of huddled instances. These subsets are collected according to two controllable thresholds. The first one is an intersection-over-union threshold $T_O$, to determine whether an instance is huddled, \emph{i.e.}, a huddled instance should have a large overlap ($> T_O$) with other instances; The second one is a proportion threshold $T_P$, to determine whether an image contain a large proportion of huddled instances, \emph{i.e.}, the number of huddled instances over the number of all instances in the image should be $> T_P$. We collect the subsets by varying $T_O$ and $T_P$ from 0.0 to 0.4 with an interval of 0.1.
\paragraph{Evaluation Metric} The standard COCO-style evaluation metric is mean Average Precision (AP) over classes. But, in a subset, some classes may only have very a few instances or even no instance, making the standard AP heavily unstable and biased. Therefore, we compute {\bf{AP over all instances}}, denoted by APoI, as the metric to evaluate the results on the subsets instead.
\paragraph{Results} We use 1$\times$ schedule to train an F-DSC and an HTC. The backbone is R-50 FPN. We report the improvements in terms of box APoI and mask APoI of DSC over HTC in Table~\ref{tab:subsets_validation}. Note that, when $T_O=-1.0$, the selected subset is the original COCO 2017 \texttt{val} set, since all instances are determined as huddled instances. F-DSC outperforms HTC by 2.4 box APoI and 2.1 mask APoI respectively on the original set. Note that, the improvements in terms of APoI are similar to the improvements in terms of AP reported in Table~\ref{tab:DSC validation}, which shows APoI is a reasonable metric. With the increase of $T_O$ and $T_P$, we observe the improvements of DSG over HTC become larger, achieving {\bf 4.2} box APoI and {\bf{4.7}} mask APoI on subsets with high proportions of heavily overlapped instances. These results evidence the advantage of DSC on detecting and segmenting huddled instances.

\begin{table}[!h]
\begin{center}
\small
\setlength{\tabcolsep}{0.26em}
\begin{tabular}{|c|c|c|c|c|c|c|}
\hline
\diagbox{$T_O$}{$T_P$} & 0.0  &   0.1 &  0.2 &   0.3 &   0.4  \\
\hline
 -1.0 & 2.4 / 2.1 & - & - & - & - \\
\hline
 0.0 & 2.5 / 2.0 & 2.5 / 2.0 & 2.6 / 2.2 & 2.7 / 2.4 & 2.5 / 2.4  \\
\hline
 0.1 & 2.5 / 2.0 & 2.4 / 2.1 & 2.6 / 2.3 & 2.8 / 2.5 & 2.8 / 2.8 \\
\hline
 0.2 & 2.5 / 2.1 & 2.5 / 2.2 & 2.7 / 2.6 & 2.9 / 2.9 & 3.0 / 3.3 \\
\hline
 0.3 & 2.4 / 2.0 & 2.6 / 2.4 & 3.0 / 2.9 & 3.7 / 3.6 & 3.6 / 3.9\\
\hline
 0.4 & 2.5 / 2.1 & 2.9 / 2.6 & 3.2 / 3.2 & 3.8 / 3.9 & {\bf 4.2} / {\bf 4.7}  \\
\hline
\end{tabular}
\end{center}
\caption{Improvements of F-DSC over HTC on subsets of COCO 2017 \texttt{val} with different proportions of huddled instances. In each table cell, we report the box APoI improvement / the mask APoI improvement on the subset selected by using the two thresholds $T_O$ and $T_P$. }
\label{tab:subsets_validation}
\end{table}

\section{Limitation}
We investigate the failure cases of our proposed method to study the limitation of our method. Since our method is shape-guided, it can be imagined that the final result quality of our method relies on informative initial mask predictions. So, bad initial mask predictions might worsen final results. To verify this hypothesis, we investigate the comparison between DSC and HTC case-by-case and find that 1) DSC performs worse on only a few cases ($\approx 10\%$), 2) the qualities of the initial mask predictions of these cases are low ($\approx 0.3$ mask mean IoU). To address this limitation, we can explore the direction of estimating mask IoUs, e.g. Mask Scoring, and re-weighting shape guidances according to the estimated IoUs.
\section{Conclusion}


In this paper, we proposed Deeply Shape-guided Cascade (DSC) for instance segmentation, which iteratively makes use of the shape guidances extracted from mask segmentation at previous stage to improve bounding box detection at current stage. Then, the improved bounding box detection results can lead to more precise mask segmentation at current stage. This forms a positive feedback loop between bounding box detection and mask segmentation across multiple stages in the cascade, establishing a bi-directional relationship between the two tasks. The experimental results on the COCO benchmark showed that DSC achieved outperforms the state-of-the-art instance segmentation cascade, HTC, by a large margin. Particularly, DSC achieved significant improvements over HTC on segmenting huddled instances.

{\small
\bibliographystyle{ieee_fullname}
\bibliography{egbib}
}

\end{document}